\documentclass{article}

\usepackage{arxiv}

\usepackage[utf8]{inputenc} 
\usepackage[T1]{fontenc}    
\usepackage{hyperref}       
\usepackage{url}            
\usepackage{booktabs}       
\usepackage{amsfonts}       
\usepackage{nicefrac}       
\usepackage{microtype}      
\usepackage{lipsum}		
\usepackage{graphicx}
\usepackage{natbib}
\usepackage{doi}
\usepackage{multirow}
\usepackage{adjustbox}
\usepackage{stfloats}
\usepackage{subfig}
\usepackage{amsmath}
\usepackage{amsthm}
\usepackage{amssymb}

\title{LeMoF: Level-guided Multimodal Fusion for Heterogeneous Clinical Data}

\author{
Jongseok Kim
\thanks{These authors contributed equally to this work.}
\\
Department of Computer Science\\
Chungbuk National University\\
South Korea\\
\texttt{kjseok@chungbuk.ac.kr}
\And
Seongae Kang
\footnotemark[1]
\\
Department of Computer Science\\
Chungbuk National University\\
South Korea\\
\texttt{kseongae@chungbuk.ac.kr}
\And
Jonghwan Shin\\
Department of Computer Science\\
Chungbuk National University\\
South Korea\\
\texttt{shinjh0218@chungbuk.ac.kr}
\And
Yuhan Lee \\
Brigham and Women's Hospital\\
Harvard Medical School\\
Boston, United States\\
\texttt{ylee21@bwh.harvard.edu}
\And
Ohyun Jo \\
Department of Computer Science\\
Chungbuk National University\\
South Korea\\
\texttt{ohyunjo@chungbuk.ac.kr}
}

\date{}

\renewcommand{\shorttitle}{LeMoF: Level-guided Multimodal Fusion for Heterogeneous Clinical Data}

\begin{document}
\maketitle

\begin{abstract}
    Multimodal clinical prediction is widely used to integrate heterogeneous data such as Electronic Health Records (EHR) and biosignals. However, existing methods tend to rely on static modality integration schemes and simple fusion strategies. As a result, they fail to fully exploit modality-specific representations.
    In this paper, we propose Level-guided Modal Fusion (LeMoF), a novel framework that selectively integrates level-guided representations within each modality. Each level refers to a representation extracted from a different layer of the encoder. LeMoF explicitly separates and learns global modality-level predictions from level-specific discriminative representations. This design enables LeMoF to achieve a balanced performance between prediction stability and discriminative capability even in heterogeneous clinical environments. Experiments on length of stay prediction using Intensive Care Unit (ICU) data demonstrate that LeMoF consistently outperforms existing state-of-the-art multimodal fusion techniques across various encoder configurations. We also confirmed that level-wise integration is a key factor in achieving robust predictive performance across various clinical conditions.
\end{abstract}

\keywords{Multimodal Learning \and Hierarchical Representation Learning \and Clinical Time-Series Modeling \and Level-guided Feature Fusion, Explainable Medical AI}

\section{Introduction}
In modern medicine, the integration of multimodal clinical information for medical diagnosis is a common practice, and extensive research has focused on combining heterogeneous data sources, modalities such as Electronic Health Records (EHRs), medical images, and text reports to support clinical decision-making. \cite{zhou2023transformer} Multi-perspective analysis of patient conditions leveraging diverse information sources leads to improved diagnostic accuracy and enables earlier diagnosis in challenging tasks such as Length of Stay (LOS) prediction. However, since EHR and Chest X-Ray (CXR) are often integrated together, they tend to be limited to specific modalities. \cite{wolf2022daft,yao2024drfuse,hayat2022medfuse,duenias2025hyperfusion,joze2020mmtm} Furthermore, many multimodal methods depend on straightforward fusion schemes, failing to explicitly model the unique characteristics and information structures inherent to each modality. \cite{duan2024deep} In this regard, established multimodal techniques have several challenges.

\subsection{Challenge 1 : A Tendency Toward Modality-Specific Dependence.}
EHR data captures rich phenotypic information about patients across time, while CXR data provides visual information practical for lesion identification. \cite{chen2024predictive,yao2024drfuse} However, since EHR and CXR represent information as static snapshots, merely the fusion of the two modalities is insufficient to reflect the progression of a patient's condition. \cite{huang2020fusion} In contrast, ECG contains high-density information that represent the electrophysiological status of the patient's heart and potential cardiovascular risk. \cite{chen2025deri} While ECG signals are highly informative and constitute one of the most important biosignals, their inherent signal characteristics limit their performance as a single modality. Therefore, fusion with EHR, which provide intricate patient information, may enhance diagnostic accuracy while capturing complementary relationships in multimodal. However, since EHR and ECG are comprised of different forms and dimensions and demonstrate different information, the characteristics of each modality must be integrated without loss to effectively reflect the discriminative information of multimodalities.

\subsection{Challenge 2 : The Necessity of Selective Integration in Multimodal Healthcare. }
Multimodal data fusion leverages heterogeneous data sources to interpret a patient’s current health status across diverse clinical contexts. Nevertheless, due to the heterogeneous and complex nature of large-scale medical data, information integration does not necessarily result in improved predictive performance. \cite{kim2023heterogeneous} EHR represents high-dimensional tabular information, while ECG consists of time-series signals that are inherently noisy. Within a multi-layered model, representations extracted at different depths exhibit diverse characteristics corresponding to different levels of abstraction. Even within each data, predictive performance may vary depending on the information being reflected. Accordingly, effective multimodal learning extends beyond simple fusion and requires selectively integrating modality-specific information while preserving their unique characteristics.

To address these challenges, we proposes a novel framework called LeMoF (Level-guided Modal Fusion), which consist of three key modules. LeMoF optimizes learning based on hierarchical representations of each modality through modality-specific training, seamlessly integrating the features of each modality. Ultimately, predictions from the two modules are stacked to generate final integrated predictions. (Fig 1.)

\begin{figure}[t]
\centering
\includegraphics[width=2.5in]{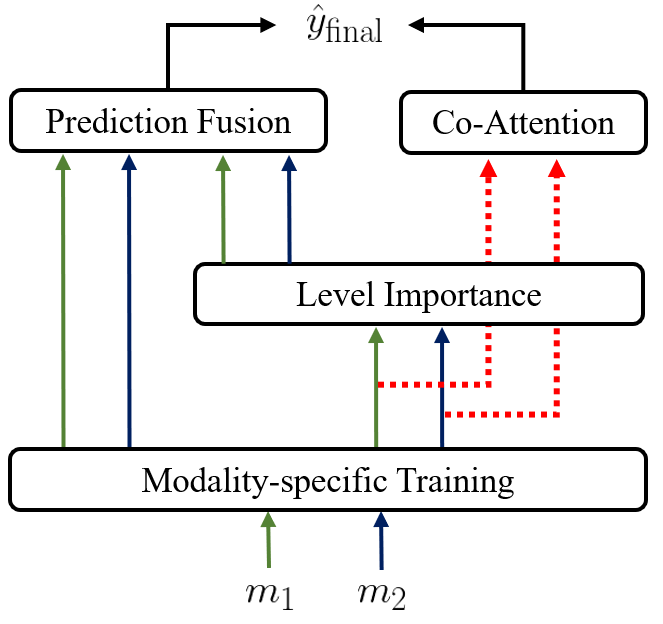}
\caption{LeMoF framework. Hierarchical representations from different encoder stages are selectively integrated via level-aware fusion.}
\label{first figure}
\end{figure}

The main contributions of this paper are as follows. We propose a novel multimodal framework that incorporates modality-aware level stacking. It ensures stable performance by hierarchically reflecting multi-level information through level importance and prediction fusion. Level-guided multimodal fusion utilizes complementary relationships between modalities as additional information, enabling it to support heterogeneous modalities. We validated our approach for various diseases, focusing on the LOS prediction task, and achieved superior performance compared to state-of-the-art fusion techniques.

\section{Related Work}
\subsection{Multi-modal learning in medical domain}
Multi-modal fusion approaches combining heterogeneous medical data modalities such as electronic health records (EHR), Electrocardiogram (ECG), and chest X-rays (CXR) have been extensively studied. In the medical field, fusion techniques have demonstrated strong performance  in tasks such as readmission prediction, disease classification and mortality prediction. [Hayat et al., 2022] conducted phenotype classification and mortality prediction using clinical time-series and CXR data.  [Wang et al., 2023] examined cardiac condition classification by extracting invariant representations from ECG signals. While ECG data is a crucial for clinical task, existing studies have focused on EHR and CXR fusion or single ECG modality.  ECG data provide continuous cardiac information, whereas CXR reflect only specific states. Consequently, ECG data contain more temporal feature, which is essential for clinical tasks.  Additionally, using ECG data alone is insufficient to fully represent comprehensive patient characteristics. Thus,  two complementary modalities are employed - EHR for diverse patient attributes and ECG for heart-related signals. [Duenias et al., 2025] proposed a HyperFusion framework that dynamically generates the weights of image analysis networks conditioned on tabular data for predicting brain age and AD classification. [Yao et al., 2024] introduced DrFuse to address missing modality and modal inconsistency. [Chen et al., 2025] Suggested the DERI approach based on multiple feature alignment and mutual reconstruction. However, these studies suffer from information loss by relying solely on compressed representations produced by the encoder. To address this issue, features are utilized at every level of each modality, and the most important level is emphasized in training. By doing so, the model preserve critical information and benefits from representations captured at various stages.

\subsection{Length of Stay prediction}
Accurate hospital length of stay (LOS) prediction is critical to effective healthcare resource management and the mitigation of intensive care unit (ICU) bed shortages. Traditional prediction methods determine hospital length of stay using patient information or ICU-specific characteristics. However, patient length of stay differs considerably across disease types. Therefore, incorporating disease-specific characteristics is essential for accurate forecasting. To address these limitation, recent studies have proposed LOS prediction models focusing on specific patient diseases. [Alsinglawi et al., 2022] proposed a Random Forest–based LOS prediction framework for lung cancer patients using electronic health records (EHR), incorporating SHAP to enhance model interpretability. Similarly, [Yang et al., 2022] estimated hospital length of stay for patients with acute ischemic stroke (AIS) using an deep learning-based architectures trained on 14 clinical tabular features. Despite their effectiveness, these approaches utilizing static clinical characteristics have inherent limitations, as they do not adequately capture the dynamic physiological states that occur during ICU stays. Moreover, in real-world ICUs where patients with diverse diseases share the same bed resources, predictive approaches focused on a single disease has limited applicability for efficient medical resource management. Therefore, a generalized LOS prediction model that is not restricted to a specific disease and can accommodate diverse patient populations is required.

\section{LeMoF : Level-guided Modal Fusion}
\begin{figure*}[t]
\centering
\includegraphics[width=6.4in]{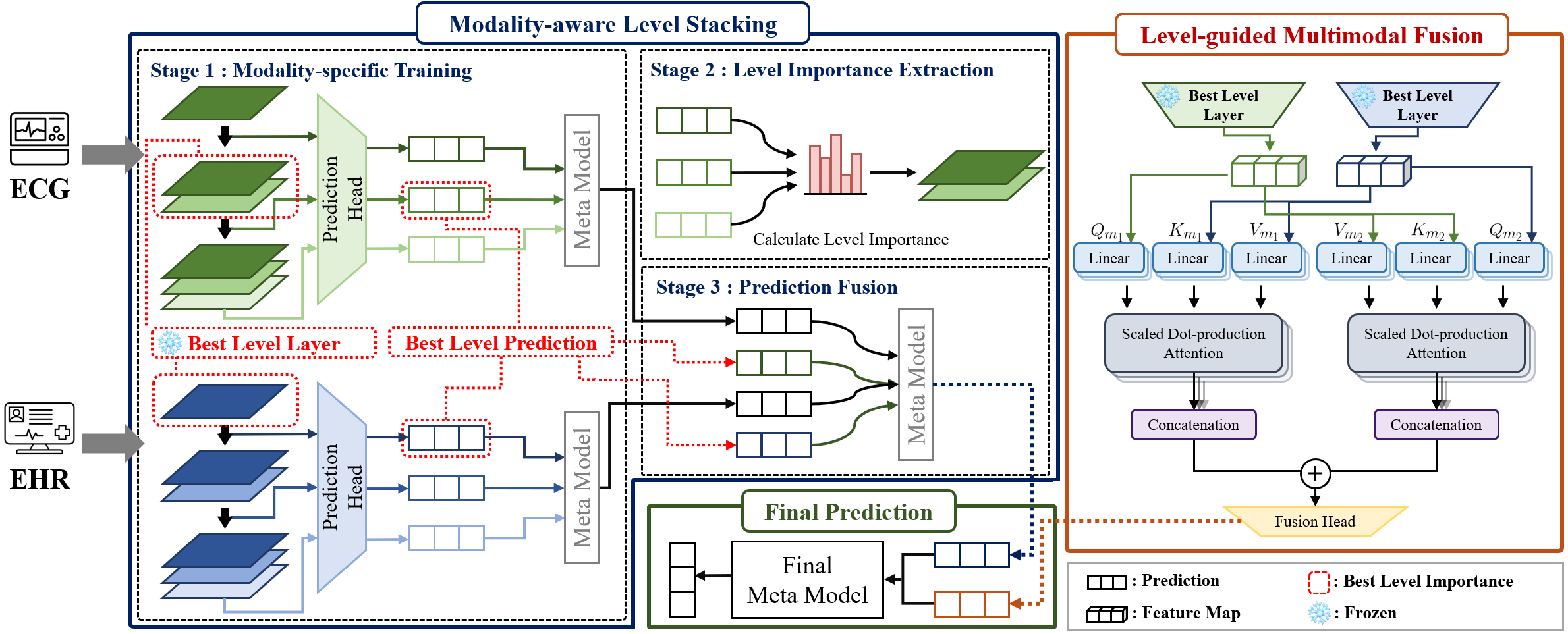}
\hfil
\caption{Detailed illustration of the proposed LeMoF framework. Hierarchical representations are extracted from each modality using a Pyramid Feature Network, followed by level-wise prediction learning and importance-based best-level selection (Module 1 $\rightarrow$ M1). The selected representations are then used for level-guided cross-modal attention (Module 2 $\rightarrow$ M2) and integrated with modality-internal predictions to produce the final output (Module 3 $\rightarrow$ M3).}
\label{main figure}
\end{figure*}

In this study, we propose LeMoF, a multimodal learning framework designed to effectively model heterogeneous ICU data. Figure \ref{main figure} provides an overview of the proposed architecture, which consists of three key modules: (1) \textit{Modality-aware Level Stacking}; (2)\textit{Level-guided Multimodal Fusion}; and (3) \textit{Integrated Decision Modeling}. The first module exploits intra-modality level information, the second enhances cross-modal interactions, and the final module integrates multi-stage predictions through stacking.

\subsection{Modality-aware Level Stacking}
ICU data consist of heterogeneous modalities with different temporal resolutions and representational characteristics, and each modality provides predictive information at distinct levels and in different forms. Learning based on a single representation level is therefore insufficient to fully capture such complex structures. To address this issue, this module is designed to explicitly extract multi-level hierarchical representations from each modality and to systematically exploit them in the prediction space. Modality-aware Level Stacking is composed of three stages: (i) Stage 1: Modality-specific Training, (ii) Stage 2: Level Importance Extraction, and (iii) Stage 3: Best-level Aggregation. Each stage is responsible for learning hierarchical representations within a modality, identifying the most informative representation levels, and integrating them from a predictive perspective.

\subsubsection{Stage 1 : Modality-specific Training}

The objective of Stage 1 is to capture predictive information within each modality across multiple representation levels. Low-level features typically encode fine-grained characteristics such as local signal patterns or variable-level physiological measurements, whereas high-level features capture more abstract and stable clinical patterns. Since these representations provide complementary information for clinical prediction, relying on a single level is insufficient. 

To this end, Stage 1 introduces level-specific prediction heads to preserve predictive signals emerging at different hierarchical depths. The resulting level-wise predictions are subsequently integrated through a meta-learner, allowing diverse latent representations within each modality to be jointly exploited. The modality set is denoted as $\mathcal{M} = \{m_1, m_2\}$, where $m_1 = \mathrm{ECG}$ and $m_2 = \mathrm{EHR}$. For each modality $m \in \mathcal{M}$, hierarchical representations are extracted using a Pyramid Feature Network (PFN) composed of level-specific transformation blocks. Formally, the representation at level $k$ is defined as:
\begin{equation}
\mathbf{h}_{m}^{(k)} = \mathcal{F}_{m}^{(k)}
\!\left(
    W_{m}^{(k)} * \mathbf{h}_{m}^{(k-1)} + \mathbf{b}_{m}^{(k)}
\right),
\quad
\mathbf{h}_{m}^{(0)} = x_m
\label{eq:1}
\end{equation}
where $\mathcal{F}_{m}^{(k)}$ denotes the $k$-th block of a modality-specific backbone network. Unlike conventional approaches that rely solely on the final-layer output, PFN explicitly exposes intermediate block outputs as hierarchical representations. In this formulation, $k = 1, 2, 3$ correspond to representations at low, intermediate, and high levels. 

Each level-wise representation is fed into a dedicated prediction head to produce a level-specific predictive signal. Accordingly, a single modality yields the following prediction vector:
\begin{equation}
P_m = [\hat{y}_m^{(1)}, \hat{y}_m^{(2)}, \hat{y}_m^{(3)}], \quad
\hat{y}_m^{(k)} = f_m^{(k)}(\mathbf{h}_m^{(k)})
\label{eq:2}
\end{equation}
where$f_m^{(k)}$ denotes the level-specific prediction head at representation level $k$, $\hat{y}_m^{(k)}$ represents the corresponding level-wise prediction, and $P_m$ is the aggregated prediction vector that collects outputs from all representation levels of modality $m$. Since level-wise predictions encode information from different representation depths, simple averaging or concatenation fails to adequately capture inter-level relationships. To overcome this limitation, a modality-specific logistic meta-learner is introduced and defined as :
\begin{equation}
G(x;\theta)=\sigma (w^{\top }x+b)
\label{eq:3}
\end{equation}
where $\theta = \{w, b\}$ denotes learnable parameters and $\sigma(\cdot)$ is the sigmoid function. Under this formulation, the aggregated prediction vector $P_m$ serves as the input to the logistic meta-learner, yielding a modality-level aggregated prediction:
\begin{equation}
\hat{y}_m^{meta} = G(P_m; \theta_m^{meta})
= \sigma(w_m^{\top} P_m + b_m).
\label{eq:4}
\end{equation}
The resulting meta prediction $\hat{y}_m^{meta}$ summarizes predictive evidence across multiple representation levels, rather than relying on a single-level output. Through this meta-learning mechanism, Stage~1 explicitly projects the hierarchical structure of modality-specific representations into the prediction space while effectively integrating complementary information across levels. Consequently, both level-wise predictions and the meta-level output provide stable and interpretable inputs for subsequent stages that model cross-modality interactions.

\subsubsection{Stage 2 : Level Importance Extraction. }

Representation levels within a modality contribute unequally to clinical prediction, and their relative importance may vary across modalities and tasks. Stage 2 aims to automatically identify the most influential representation level within each modality.

The contribution of each level-wise prediction $\hat{y}_m^{(k)}$ to the output of the meta model $G(\cdot)$ is quantitatively evaluated. Specifically, the importance function $Imp(\cdot)$ is defined based on a Shapley value formulation, which measures the marginal contribution of a prediction by averaging its effect over all possible subsets of the remaining predictions. In this work, $Imp(\cdot)$ is implemented using SHAP. Accordingly, the importance score for level $k$ is computed as :
\begin{equation}
\begin{aligned}
\phi_m^{(k)}
&= \text{Imp}(\hat{y}_m^{(k)} \mid G) \\
&=
\mathbb{E}_{S \subseteq \mathbf{P}_{m} \setminus \{\hat{y}_m^{(k)}\}}
\left[
    G(S \cup \{\hat{y}_m^{(k)}\})
    -
    G(S)
\right]
\end{aligned}
\label{eq:5}
\end{equation}
where $\phi_m^{(k)}$ represents the average marginal contribution of the level-$k$ prediction to the final meta-model output, and $S$ denotes a subset of level-wise predictions excluding $\hat{y}_m^{(k)}$. By construction, a larger magnitude of $\phi_m^{(k)}$ indicates a stronger influence of the corresponding representation level on the final prediction.

Based on these importance scores, the most influential representation level, referred to as the \textit{best-level}, is selected as
\begin{equation}
k_m^{*}
=
\arg\max_{k}
\left|
    \phi_m^{(k)}
\right|,
\label{eq:6}
\end{equation}
\begin{equation}
\hat{y}_m^{(best)} = \hat{y}_m^{(k_m^{*})}.
\label{eq:7}
\end{equation}
The resulting best-level prediction $\hat{y}_m^{best}$ serves as a key anchor for Module 2, enabling the model to focus on the most informative representation while reducing redundancy across levels. This selection mechanism further improves the interpretability and stability of the proposed framework.

\subsubsection{Stage 3 : Best-level Aggregation}

Stage 3 aggregates modality-internal predictions into a unified predictive signal. Specifically, it integrates the modality-level meta prediction obtained in Stage 1 with the best-level prediction identified in Stage 2 through a second-stage meta-learning process. This aggregation allows complementary predictive information captured at different levels to be jointly exploited.

The predictions from modalities $m_1$ and $m_2$ are concatenated to form an input vector $z$, which is subsequently fed into the logistic predictor defined in the previous stages:
\begin{equation}
\hat{y}_{\Omega_1} = G(z;\theta_{\Omega_1}),
\quad
z =
\left[
\hat{y}_{m_1}^{meta},\;
\hat{y}_{m_1}^{best},\;
\hat{y}_{m_2}^{meta},\;
\hat{y}_{m_2}^{best}
\right].
\label{eq:8}
\end{equation}
where $G(\cdot)$ follows the same logistic formulation as in Stage 1 and Stage 2, while employing a distinct set of parameters $\theta_{\Omega_1}$. The resulting prediction $\hat{y}_{\Omega_1}$ provides a compact summary of the hierarchical predictive structure within each modality and serves as a high-quality input for the subsequent cross-modal fusion process in Module 2.

\subsection{Level-guided Multimodal Fusion}
While Module 1 normalizes and summarizes the hierarchical predictive structure within each modality, Module 2 is designed to learn refined interactions across modalities based on this internal structure. ICU data inherently contain complementary clinical information distributed across different modalities, making it essential to effectively model semantic dependencies between them for reliable prediction. To this end, Module 2 adopts the best-level representations identified in Module 1 as key anchors and models cross-modal interactions through an attention-based mechanism. The selected best-level representations for modalities $m_1$ and $m_2$ are defined as:
\begin{equation}
\mathbf{h}_{m_1}^{(best)} = \mathbf{h}_{m_1}^{(k_{m_1}^{*})}, \quad\mathbf{h}_{m_2}^{(best)} = \mathbf{h}_{m_2}^{(k_{m_2}^{*})}.
\label{eq:9}
\end{equation}
These best-level representations are obtained from Stage 2 of Module 1 based on the defined importance measure and correspond to the most clinically informative hierarchical features within each modality. Module 2 leverages these representations to perform cross-modal attention, thereby facilitating efficient and targeted information exchange between the two modalities.

Specifically, each modality computes Query, Key, and Value projections to attend to the best-level representation of the other modality:
\begin{equation}
\begin{aligned}
Q_{m_i} &= W_{Q}^{m_i} \mathbf{h}_{m_i}^{*}, \;
K_{m_j} = W_{K}^{m_j} \mathbf{h}_{m_j}^{*}, \;
V_{m_j} = W_{V}^{m_j} \mathbf{h}_{m_j}^{*}, \\
&\qquad \forall (m_i, m_j) \in \{ (m_1, m_2), (m_2, m_1) \}.
\end{aligned}
\label{eq:10}
\end{equation}
The first case corresponds to modality $m_1$ attending to modality $m_2$, while the second represents the reverse direction. Based on these projections, the cross-modal attention outputs are computed as:
\begin{equation}
\text{Attn}_{{m_i} \leftarrow {m_j}}
=
{softmax}\!\left(
\frac{Q_{m_i} K_{m_j}^{\top}}{\sqrt{d}}
\right) V_{m_j}.
\label{eq:11}
\end{equation}
This cross-modal attention structure enables each modality to selectively incorporate only the most informative signals from the other modality. In particular, since attention is computed using the best-level representations selected in Module~1, the cross-modal fusion process naturally suppresses redundant or less informative hierarchical features and focuses on representations with the highest clinical relevance.

The attention outputs from both directions are concatenated to form a fused cross-modal representation:
\begin{equation}
\hat{y}_{\Omega_2} = G(c;\theta_{\Omega_2}),
\quad
c = \left [\text{Attn}_{m_1 \leftarrow m_2}, \; \text{Attn}_{m_2 \leftarrow m_1}  \right ].
\label{eq:12}
\end{equation}
where $c$ represents a complementary cross-modal embedding that jointly captures the interactions between modalities $m_1$ and $m_2$ and serves as an informative feature basis for subsequent prediction. By design, Module~2 structurally regulates cross-modal interactions while mitigating unnecessary representation redundancy. As a result, it effectively captures complex dependencies present in multimodal ICU data and provides a well-conditioned input for the final prediction stage in Module 3.

\subsection{Integrated Decision Modeling}
Module 3 produces the final prediction by integrating the outputs obtained from the preceding modules. Specifically, it combines the modality-level aggregated prediction $\hat{y}_{\Omega_1}$ derived from Module 1 with the multimodal prediction $\hat{y}_{\Omega_2}$ generated by Module 2, thereby jointly exploiting predictive evidence from both modality-internal representations and cross-modal interactions.

The two predictions are concatenated into a unified input vector, and the final output is computed using the same logistic prediction function $G(\cdot)$ employed throughout the framework:
\begin{equation}
\hat{y}_{\text{final}} = G(\mathbf{u}; \theta_{\text{final}}), \quad
\mathbf{u} =
\left[
\hat{y}_{\Omega_1},\;
\hat{y}_{\Omega_2}
\right].
\label{eq:13}
\end{equation}
where $G(\cdot)$ follows the logistic formulation defined in the previous modules, while introducing a dedicated set of learnable parameters $\theta_{\text{final}}$ for the final integration step. By consolidating predictive signals summarized at different stages of the framework, Module~3 yields a coherent and well-conditioned final prediction. This final aggregation step completes the end-to-end inference process of LeMoF by reconciling hierarchical aggregation and cross-modal fusion within a single decision function.

\section{Experiment}
\subsection{Experimental Setting}
To demonstrate the performance of our proposed framework, we conduct experiments on the MIMIC-IV and MIMIC-IV-ECG datasets \cite{johnson2023mimic,gow2023mimic}. MIMIC-IV provides clinical records of patients admitted to ICU and emergency departments at the Beth Israel Deaconess Medical Center in Boston. MIMIC-IV-ECG consists of  800,035 ECG recordings sampled at 500 Hz for 10 seconds from 161,352 unique patients. 
Using these datasets, the performance of LeMoF is evaluated on a length-of-stay (LoS) prediction task.
In LeMoF, the core component, Module 1, is denoted as M1, while Module 2 is denoted as M2.

\begin{table*}[t]
\centering
\renewcommand{\arraystretch}{1.3}
\begin{adjustbox}{max width=\textwidth}
\begin{tabular}{lllcclcclcclcclcclcc}
\hline\hline
\multicolumn{2}{l}{\textbf{Model}}    &  & \multicolumn{2}{c}{\begin{tabular}[c]{@{}c@{}}DAFT\\ \cite{wolf2022daft}\end{tabular}} &  & \multicolumn{2}{c}{\begin{tabular}[c]{@{}c@{}}MedFuse\\ \cite{hayat2022medfuse}\end{tabular}} &  & \multicolumn{2}{c}{\begin{tabular}[c]{@{}c@{}}DrFuse\\ \cite{yao2024drfuse}\end{tabular}} &  & \multicolumn{2}{c}{\begin{tabular}[c]{@{}c@{}}MMTM\\ \cite{joze2020mmtm}\end{tabular}} &  & \multicolumn{2}{c}{\begin{tabular}[c]{@{}c@{}}HyperFusion\\ \cite{duenias2025hyperfusion}\end{tabular}} &  & \multicolumn{2}{c}{\textbf{\begin{tabular}[c]{@{}c@{}}LeMoF\\ (Ours)\end{tabular}}} \\ \cline{1-2} \cline{4-5} \cline{7-8} \cline{10-11} \cline{13-14} \cline{16-17} \cline{19-20} 
ECG                       & EHR       &  & ACC                                       & AUROC                                    &  & ACC                                          & AUROC                                        &  & ACC                                        & AUROC                                      &  & ACC                                       & AUROC                                    &  & ACC                                               & AUROC                                             &  & ACC                                      & AUROC                                    \\ \hline\hline
\multirow{5}{*}{Baseline} & Baseline  &  & \underline{0.609}                            & 0.649                                    &  & 0.602                                        & 0.645                                        &  & 0.597                                      & 0.649                                      &  & 0.593                                     & \underline{0.650}                           &  & 0.581                                             & 0.636                                             &  & \textbf{0.667}                           & \textbf{0.655}                           \\
                          & FT-Trans. &  & 0.505                                     & 0.622                                    &  & \underline{0.560}                               & 0.627                                        &  & 0.525                                      & 0.619                                      &  & 0.525                                     & 0.616                                    &  & 0.537                                             & \underline{0.635}                                    &  & \textbf{0.689}                           & \textbf{0.691}                           \\
                          & TPC       &  & \underline{0.646}                            & \underline{0.695}                                    &  & 0.625                                        & 0.689                                        &  & 0.641                                      & 0.689                                      &  & 0.644                                     & \textbf{0.695}                           &  & 0.643                                             & 0.693                                             &  & \textbf{0.680}                           & 0.680                                    \\
                          & TabNet    &  & 0.613                                     & 0.587                                    &  & 0.608                                        & 0.592                                        &  & 0.642                                      & 0.605                                      &  & \underline{0.644}                            & \underline{0.609}                           &  & 0.637                                             & 0.601                                             &  & \textbf{0.656}                           & \textbf{0.611}                           \\
                          & TabTrans. &  & 0.587                                     & 0.638                                    &  & 0.575                                        & 0.619                                        &  & 0.592                                      & 0.638                                      &  & \underline{0.592}                            & \underline{0.641}                           &  & 0.577                                             & 0.631                                             &  & \textbf{0.666}                           & \textbf{0.648}                           \\ \hline
\multirow{5}{*}{LSTM}     & Baseline  &  & 0.585                                     & 0.637                                    &  & \underline{0.606}                               & 0.646                                        &  & 0.588                                      & 0.639                                      &  & 0.602                                     & \underline{0.646}                           &  & 0.600                                             & 0.643                                             &  & \textbf{0.668}                           & \textbf{0.652}                           \\
                          & FT-Trans. &  & \underline{0.573}                            & 0.578                                    &  & 0.551                                        & \underline{0.629}                               &  & 0.546                                      & 0.608                                      &  & 0.532                                     & 0.611                                    &  & 0.509                                             & 0.620                                             &  & \textbf{0.687}                           & \textbf{0.692}                           \\
                          & TPC       &  & 0.637                                     & 0.689                                    &  & 0.635                                        & \underline{0.690}                               &  & 0.636                                      & 0.689                                      &  & \underline{0.649}                            & \textbf{0.694}                           &  & 0.643                                             & 0.689                                             &  & \textbf{0.681}                           & 0.684                                    \\
                          & TabNet    &  & 0.644                                     & 0.590                                    &  & 0.616                                        & 0.602                                        &  & \underline{0.651}                             & \underline{0.611}                             &  & 0.643                                     & 0.590                                    &  & 0.640                                             & 0.578                                             &  & \textbf{0.653}                           & \textbf{0.613}                           \\
                          & TabTrans. &  & 0.573                                     & 0.623                                    &  & 0.561                                        & 0.614                                        &  & 0.567                                      & 0.637                                      &  & \underline{0.592}                            & \underline{0.640}                           &  & 0.572                                             & 0.629                                             &  & \textbf{0.665}                           & \textbf{0.647}                           \\ \hline
\multirow{5}{*}{ResNet}   & Baseline  &  & \underline{0.606}                            & 0.642                                    &  & 0.586                                        & \underline{0.645}                               &  & 0.577                                      & 0.638                                      &  & 0.583                                     & 0.640                                    &  & 0.542                                             & 0.631                                             &  & \textbf{0.667}                           & \textbf{0.653}                           \\
                          & FT-Trans. &  & \underline{0.546}                            & 0.614                                    &  & 0.509                                        & 0.624                                        &  & 0.512                                      & 0.596                                      &  & 0.529                                     & 0.620                                    &  & 0.523                                             & \underline{0.633}                                    &  & \textbf{0.688}                           & \textbf{0.689}                           \\
                          & TPC       &  & 0.647                                     & 0.691                                    &  & 0.645                                        & 0.691                                        &  & 0.618                                      & 0.691                                      &  & 0.636                                     & \underline{0.693}                           &  & \underline{0.651}                                    & \textbf{0.694}                                    &  & \textbf{0.683}                           & 0.684                                    \\
                          & TabNet    &  & 0.610                                     & 0.582                                    &  & 0.615                                        & 0.595                                        &  & 0.468                                      & 0.571                                      &  & \underline{0.619}                            & \underline{0.599}                           &  & 0.517                                             & 0.536                                             &  & \textbf{0.652}                           & \textbf{0.604}                           \\
                          & TabTrans. &  & 0.587                                     & \underline{0.644}                           &  & 0.568                                        & 0.613                                        &  & 0.565                                      & 0.637                                      &  & 0.583                                     & 0.640                                    &  & \underline{0.588}                                    & 0.635                                             &  & \textbf{0.667}                           & \textbf{0.648}                           \\ \hline
\multirow{5}{*}{WaveNet}  & Baseline  &  & 0.592                                     & \underline{0.649}                           &  & 0.594                                        & 0.646                                        &  & \underline{0.602}                             & 0.641                                      &  & 0.594                                     & 0.646                                    &  & 0.558                                             & 0.594                                             &  & \textbf{0.670}                           & \textbf{0.657}                           \\
                          & FT-Trans. &  & 0.516                                     & 0.618                                    &  & 0.546                                        & 0.635                                        &  & 0.541                                      & 0.630                                      &  & 0.535                                     & 0.619                                    &  & \underline{0.548}                                    & \underline{0.645}                                    &  & \textbf{0.689}                           & \textbf{0.692}                           \\
                          & TPC       &  & 0.631                                     & \underline{0.696}                           &  & 0.643                                        & 0.694                                        &  & 0.641                                      & 0.692                                      &  & \underline{0.647}                            & \textbf{0.697}                           &  & 0.635                                             & 0.683                                             &  & \textbf{0.680}                           & 0.681                                    \\
                          & TabNet    &  & 0.643                                     & 0.580                                    &  & \underline{0.627}                               & \underline{0.606}                               &  & 0.647                                      & 0.602                                      &  & 0.641                                     & 0.596                                    &  & 0.641                                             & 0.575                                             &  & \textbf{0.653}                           & \textbf{0.618}                           \\
                          & TabTrans. &  & 0.579                                     & \underline{0.642}                           &  & 0.561                                        & 0.625                                        &  & 0.574                                      & 0.635                                      &  & \underline{0.584}                            & 0.639                                    &  & 0.578                                             & 0.621                                             &  & \textbf{0.666}                           & \textbf{0.652}                           \\ \hline
\multicolumn{2}{c}{Rank (Avg)}        &  & 3.70                                      & 4.05                                     &  & 4.40                                         & 3.60                                         &  & 4.30                                       & 4.15                                       &  & 3.35                                      & 2.85                                     &  & 4.25                                              & 4.30                                              &  & \textbf{1.00}                            & \textbf{2.00}                            \\ \hline\hline
\end{tabular}
\end{adjustbox}
\caption{Performance comparison of multimodal fusion methods across different ECG–EHR encoder combinations. Accuracy (ACC) and AUROC are reported. Bold and underlined values indicate the best and second-best results, respectively. Rank (Avg) denotes the average ranking across all settings.}
\end{table*} 

\subsection{Holistic Performance}
\subsubsection{Holistic Evaluation of LoS Prediction}
Typically, for the task of predicting the length of hospital stay, validation on all patients is difficult, requiring separate validation by disease. However, in real-world clinical settings, stratifying all patients by disease is challenging; therefore, evaluation on the entire intensive care unit population is also essential \cite{alsinglawi2022explainable}. 
Accordingly, Table 1 summarizes the performance comparison between LeMoF and various existing multimodal fusion methods for the Length of Stay (LoS) prediction task. 
Under identical model architectures and experimental settings, LeMoF consistently achieves higher accuracy and AUROC than existing multimodal fusion techniques across nearly all encoder combinations. These results indicate that LeMoF can be applied in a model-agnostic manner, regardless of the specific encoder types employed. In particular, LeMoF attains the highest average rank across all experimental settings. These findings suggest that LeMoF effectively performs multimodal fusion while mitigating modality information loss compared to existing methods. These results indicate that LeMoF effectively mitigates information loss across modalities while maintaining stable performance across various backbone architectures.
Furthermore, LeMoF demonstrates strong performance when paired with modality-specific encoders, such as WaveNet for ECG and FT-Transformer for EHR. These results premise that the proposed framework has the potential to achieve further performance gains as modality-specific representation learning techniques continue to advance.
These experimental results validate the effectiveness of LeMoF as a framework that provides reliable and consistent multimodal performance for the LoS prediction task across the entire patient population.

\subsubsection{Ablation Study of Key Modules}
Figure 3 presents the results of an ablation study conducted on the key modules of LeMoF to clearly investigate the sources of its performance improvements. When prediction is performed using only a single modality, particularly for ECG (Single), both subfigures (b) and (c) exhibit substantially lower AUROC and F1-score. This indicates that accurate prediction is challenging due to the inherent complexity of ECG signal characteristics and the absence of sufficient clinical context regarding the patient. 
In contrast, the progressive improvement observed across the M1 and M2 variants demonstrates that structured clinical data effectively complements information that is difficult to capture from biosignal data alone. 
Additionally, a substantial improvement in F1-score performance is observed for M2.
This indicates that the best-level layers selected in M1 a complementary relationship, explicitly capturing meaningful information.
Hence, LeMoF consistently achieves strong AUROC and F1-score performance, which are critical evaluation metrics in relatively imbalanced medical datasets. These results demonstrate that LeMoF provides reliable and generalizable performance for predicting hospital length of stay. 
Furthermore, these results indicate that the individual modules comprising LeMoF operate in a complementary manner, collectively contributing to the overall performance improvement.

\begin{figure*}[t]
\centering
\subfloat[]{\includegraphics[width=2.0in]{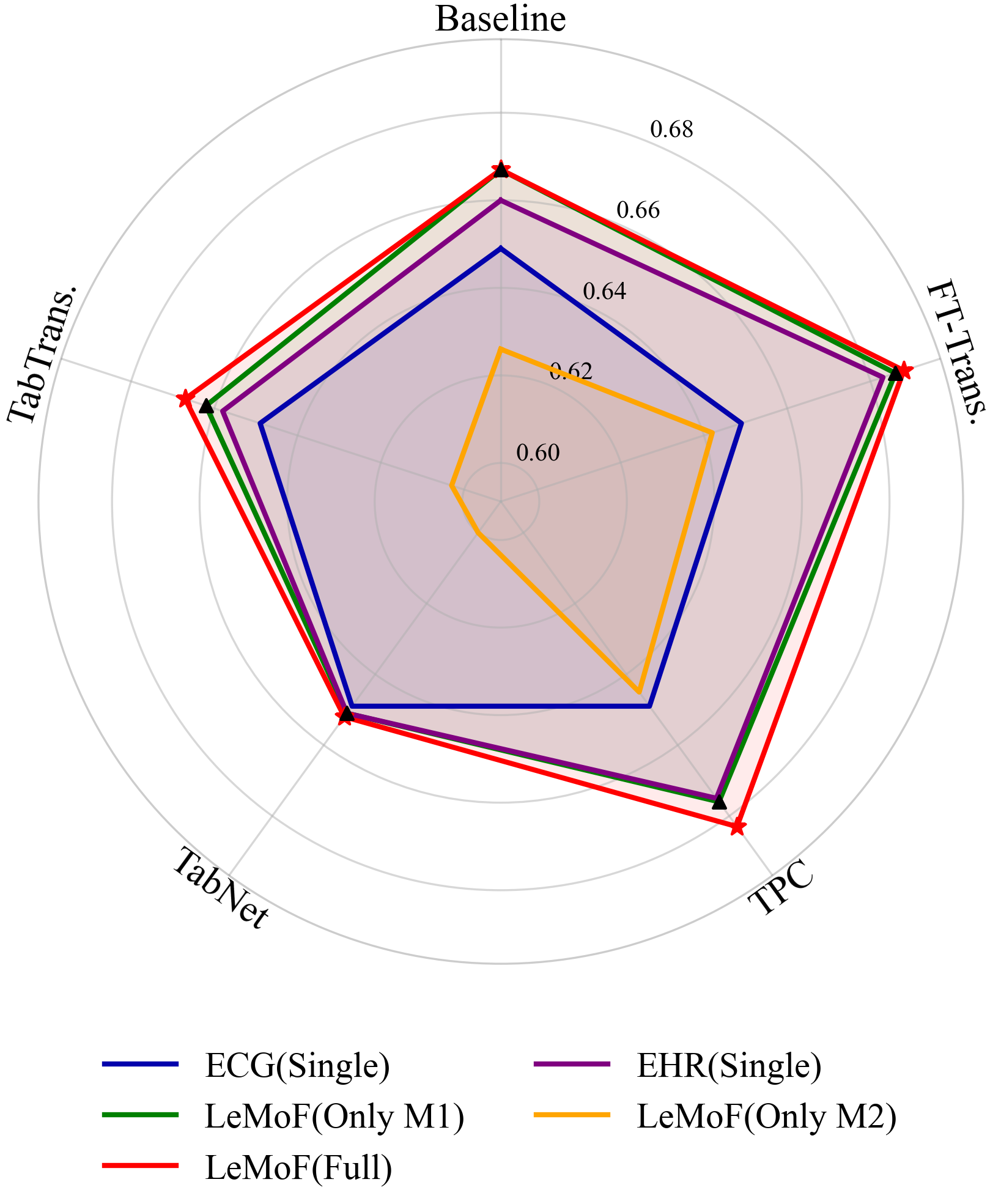}
\label{result_figure 1}}
\hfil
\subfloat[]{\includegraphics[width=2.0in]{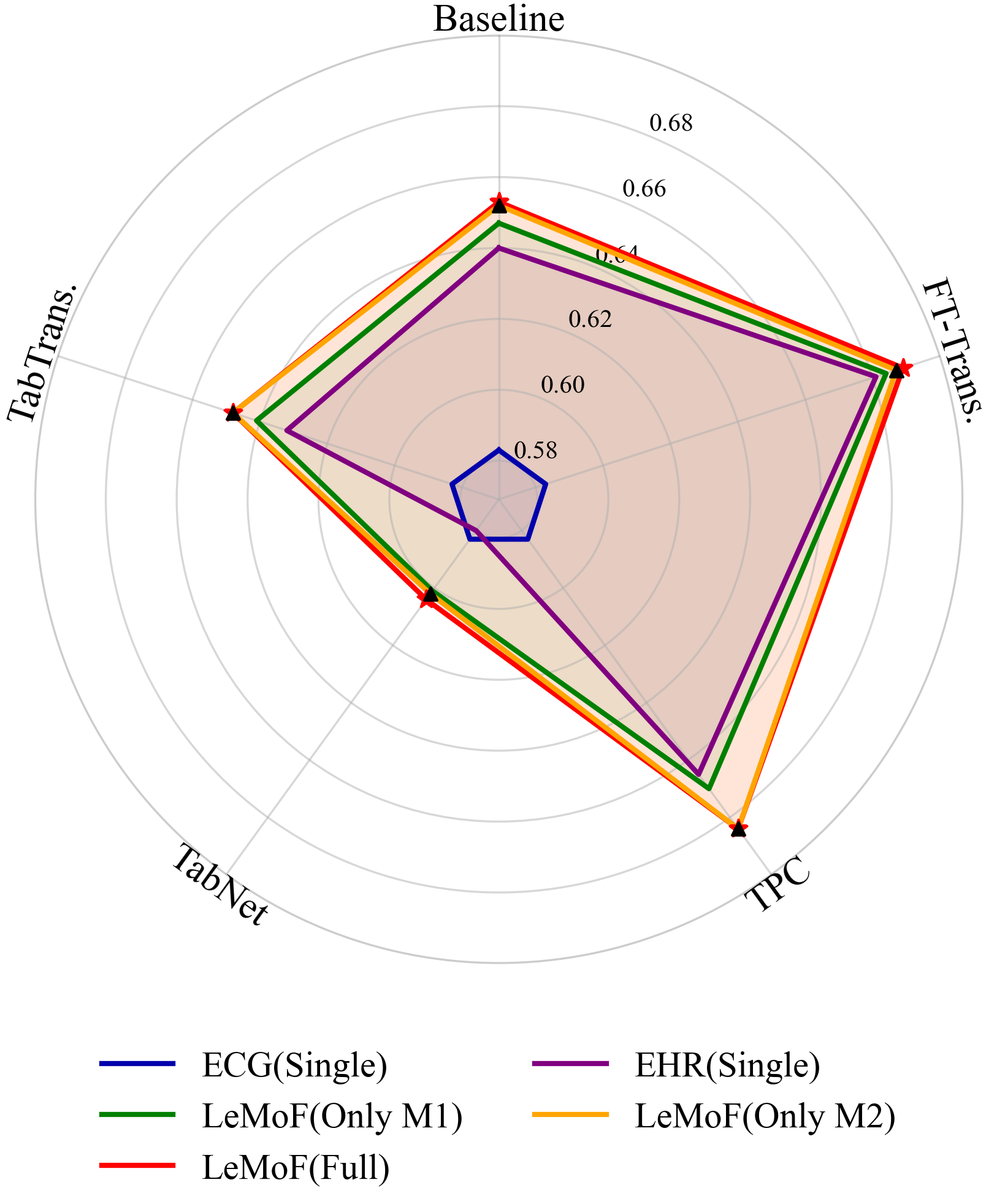}%
\label{result_figure 2}}
\hfil
\subfloat[]{\includegraphics[width=2.0in]{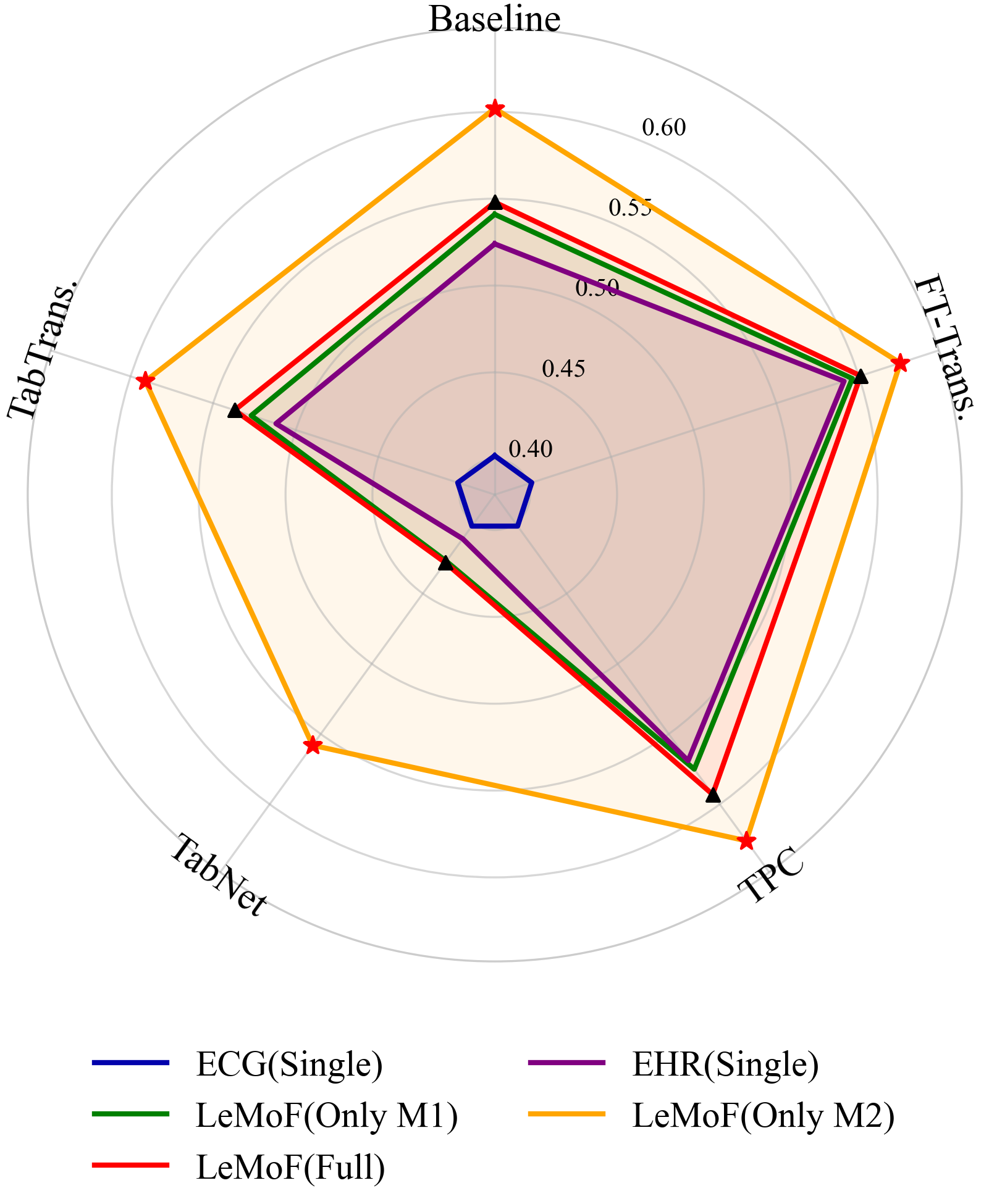}%
\label{result_figure 3}}
\hfil
\caption{Ablation study of LeMoF components under different EHR backbones, with the ECG encoder fixed to ResNet. Performance is reported in terms of (a) Accuracy, (b) AUROC, and (c) F1-score for single-modality models and LeMoF variants (Only M1, Only M2, and Full).}
\label{ablation study}
\end{figure*}
\begin{table*}[t]
\centering
\renewcommand{\arraystretch}{1.2}
\begin{adjustbox}{max width=\textwidth}
\begin{tabular}{llcccccccccccccc}
\hline\hline
\textbf{Model}                          &  & \multicolumn{2}{c}{\begin{tabular}[c]{@{}c@{}}ECG\\ (Single)\end{tabular}} &  & \multicolumn{2}{c}{\begin{tabular}[c]{@{}c@{}}EHR\\ (Single)\end{tabular}} &  & \multicolumn{2}{c}{\begin{tabular}[c]{@{}c@{}}LeMoF\\ (Only M1)\end{tabular}} &  & \multicolumn{2}{c}{\begin{tabular}[c]{@{}c@{}}LeMoF\\ (Only M2)\end{tabular}} &  & \multicolumn{2}{c}{\textbf{\begin{tabular}[c]{@{}c@{}}LeMoF\\ (Full)\end{tabular}}} \\ \cline{1-1} \cline{3-4} \cline{6-7} \cline{9-10} \cline{12-13} \cline{15-16} 
\textbf{Target}                         &  & ACC                                  & AUROC                               &  & ACC                                  & AUROC                               &  & ACC                                   & AUROC                                 &  & ACC                             & AUROC                                       &  & ACC                                     & AUROC                                     \\ \hline\hline
\textbf{Diabetes}                       &  & 0.648                                & 0.557                               &  & 0.678                                & 0.679                               &  & \underline{0.679}                     & 0.679                                 &  & 0.638                           & \underline{0.684}                           &  & \textbf{0.679}                          & \textbf{0.686}                            \\
\textbf{Lung Cancer}                    &  & 0.601                                & 0.535                               &  & 0.601                                & 0.546                               &  & \underline{0.603}                                 & 0.550                                 &  & 0.562                           & \textbf{0.557}                              &  & \textbf{0.603}                          & \underline{0.551}                         \\
\textbf{Sepsis}                         &  & 0.546                                & 0.550                               &  & 0.621                                & 0.673                               &  & \underline{0.622}                     & \underline{0.673}                     &  & 0.615                           & 0.673                                       &  & \textbf{0.627}                          & \textbf{0.677}                            \\
\textbf{Acute Myocardial Infarction}    &  & 0.640                                & 0.624                               &  & 0.662                                & 0.676                               &  & \underline{0.669}                     & \underline{0.682}                     &  & 0.639                           & 0.681                                       &  & \textbf{0.670}                          & \textbf{0.683}                            \\
\textbf{Dysrhythmia}                    &  & 0.604                       & 0.572                               &  & \textbf{0.646}                       & 0.660                               &  & \underline{0.645}                     & 0.663                                 &  & 0.624                           & \underline{0.663}                           &  & 0.645                                   & \textbf{0.665}                            \\
\textbf{Electrolyte}                    &  & 0.551                                & 0.558                               &  & 0.617                                & 0.660                               &  & \underline{0.623}                     & 0.664                                 &  & 0.617                           & \underline{0.665}                           &  & \textbf{0.623}                          & \textbf{0.668}                            \\
\textbf{Chronic Kidney Disease}         &  & 0.618                                & 0.544                               &  & 0.636                                & 0.637                               &  & \underline{0.636}                     & \underline{0.641}                     &  & 0.608                           & 0.635                                       &  & \textbf{0.637}                          & \textbf{0.641}                            \\
\textbf{Heart Failure}                  &  & 0.590                                & 0.531                      &  & 0.638                                & \textbf{0.656}                      &  & \underline{0.640}                     & 0.655                                 &  & 0.626                           & 0.653                                       &  & \textbf{0.641}                          & \underline{0.656}                         \\
\textbf{Pneumonia}                      &  & 0.558                                & 0.564                               &  & 0.602                                & 0.639                               &  & \underline{0.604}                     & \underline{0.646}                     &  & 0.602                           & 0.642                                       &  & \textbf{0.606}                          & \textbf{0.646}                            \\
\textbf{Kidney Failure}                 &  & 0.561                                & 0.565                               &  & 0.573                                & 0.636                               &  & \underline{0.567}                     & \underline{0.640}                     &  & 0.592                           & 0.637                                       &  & \textbf{0.665}                          & \textbf{0.641}                            \\ \hline
\multicolumn{1}{c}{\textbf{Rank (Avg)}} &  & 4.5                                  & 5.0                                 &  & 2.8                                  & 3.5                                 &  & \underline{1.9}                       & \underline{2.6}                       &  & 4.4                             & 2.7                                         &  & \textbf{1.2}                            & \textbf{1.2}                              \\ \hline\hline
\end{tabular}
\end{adjustbox}
\caption{Disease-wise Length of Stay prediction performance for 10 representative diseases using WaveNet as the ECG encoder and FT-Transformer as the EHR encoder. Accuracy (ACC) and AUROC are reported for single-modality models and LeMoF variants (Only M1, Only M2, and Full).}
\end{table*}

\subsection{Disease-specific Performance}
\subsubsection{Disease-specific LoS Prediction Performance}
Based on the performance validation results obtained from all intensive care unit patients, an ablation study on patients with specific diseases is further conducted to assess the versatility of LeMoF across diverse clinical conditions. 
Tables 2 and 3 present the validation results for patients with 10 representative diseases, using WaveNet as the ECG encoder and FT-Transformer as the EHR encoder, which achieved the best overall performance in Table 1.
Overall, LeMoF maintained stable and competitive performance across a wide range of disease categories. This suggests that the model exhibits robust generalization capability despite substantial heterogeneity in disease-specific predictive characteristics.

Notably, M1 consistently enhances prediction accuracy, while both M1 and M2 contribute to improvements in AUROC, thereby strengthening global discriminative performance across disease categories.

By effectively integrating these complementary strengths, LeMoF (Full) achieves superior overall performance in terms of both accuracy and AUROC. 

These findings suggest that LeMoF effectively learns informative representations at multi-levels within each modality and selectively integrates them, thereby flexibly accommodating diverse disease-specific prediction patterns.

\begin{table}[t]
\centering
\renewcommand{\arraystretch}{1.0}
\begin{adjustbox}{max width=\textwidth}
\centering
\begin{tabular}{lccccc}
\hline\hline
\textbf{Target}                         & \begin{tabular}[c]{@{}c@{}}ECG\\ (Single)\end{tabular} & \begin{tabular}[c]{@{}c@{}}EHR\\ (Single)\end{tabular} & \begin{tabular}[c]{@{}c@{}}LeMoF\\ (Only M1)\end{tabular} & \begin{tabular}[c]{@{}c@{}}LeMoF\\ (Only M2)\end{tabular} & \textbf{\begin{tabular}[c]{@{}c@{}}LeMoF\\ (Full)\end{tabular}} \\ \hline\hline
\textbf{Diabetes}                       & 0.394                                                  & 0.553                                                  & 0.553                                                     & \textbf{0.624}                                            & \underline{0.561}                                               \\
\textbf{Lung Cancer}                    & 0.382                                                  & 0.376                                                  & 0.379                                                     & \textbf{0.539}                                            & \underline{0.382}                                               \\
\textbf{Sepsis}                         & 0.471                                                  & 0.617                                                  & \underline{0.617}                                         & 0.613                                                     & \textbf{0.624}                                                  \\
\textbf{Acute Myocardial Infarction}    & 0.425                                                  & 0.588                                                  & 0.596                                                     & \textbf{0.622}                                            & \underline{0.600}                                               \\
\textbf{Dysrhythmia}                    & 0.430                                                  & 0.590                                                  & 0.591                                                     & \textbf{0.614}                                            & \underline{0.594}                                               \\
\textbf{Electrolyte}                    & 0.510                                                  & 0.609                                                  & 0.615                                                     & \underline{0.615}                                         & \textbf{0.616}                                                  \\
\textbf{Chronic Kidney Disease}         & 0.398                                                  & 0.527                                                  & 0.528                                                     & \textbf{0.590}                                            & \underline{0.537}                                               \\
\textbf{Heart Failure}                  & 0.378                                                  & 0.578                                                  & 0.582                                                     & \textbf{0.616}                                            & \underline{0.586}                                               \\
\textbf{Pneumonia}                      & 0.435                                                  & 0.582                                                  & 0.584                                                     & \textbf{0.600}                                            & \underline{0.588}                                               \\
\textbf{Kidney Failure}                 & 0.492                                                  & 0.591                                                  & 0.597                                                     & \underline{0.598}                                         & \textbf{0.599}                                                  \\ \hline
\multicolumn{1}{c}{\textbf{Rank (Avg)}} & 4.8                                                    & 4.0                                                    & 2.9                                                       & \textbf{1.6}                                              & \underline{1.7}                                                 \\ \hline\hline
\end{tabular}
\end{adjustbox}
\label{tab:complexity}
\caption{Disease-wise F1-score performance for 10 representative diseases using WaveNet as the ECG encoder and FT-Transformer as the EHR encoder. Results are reported for single-modality models and LeMoF variants (Only M1, Only M2, and Full).}
\end{table}

\subsubsection{Robustness of Level-guided Multimodal Fusion}
Table 3 reports the F1-score performance across disease categories, which is an important metric for evaluating model robustness in clinically realistic settings with class imbalance.
Consistent with the results presented in Figure 3, LeMoF (Only M2) achieves a substantial improvement in F1-score compared to single-modality models. 
These findings indicate that M2 performs a critical role in enhancing discriminative performance for challenging-to-predict cases across diverse disease categories. Furthermore, whereas M1 primarily contributes to improvements in overall prediction accuracy, M2 selectively incorporates representation levels with high predictive relevance. Thus, M2 refines the decision boundary and effectively improves the balance between precision and recall. 
Although LeMoF (Only M2) exhibits strong discriminative capability, LeMoF (Full) jointly leverages global modality-level predictions and level-specific representations, resulting in more stable performance across diseases. This complementary integration establishes a balanced trade-off between robustness and discrimination. Consequently, these results demonstrate that LeMoF effectively accommodates disease-specific prediction patterns and enhances model robustness in real-world clinical prediction environments.

\section{Conclusion}
In this study, we propose LeMoF, a level-guided multimodal fusion framework that selectively integrates hierarchical representations within each modality. LeMoF's core principle is a representation-level fusion strategy that explicitly decomposes modality-internal representations and combines them through a multi-stage process. This framework facilitates the effective utilization of informative representations from modalities with heterogeneous characteristics.
In particular, level-specific integration was essential in enhancing model robustness in clinical environments characterized by class imbalance and disease heterogeneity.
Our results suggest that LeMoF provides a generalizable foundation for diverse multimodal medical applications through its selective integration mechanism at the architectural level.

\bibliographystyle{unsrtnat}
\bibliography{references}  






\end{document}